\documentclass{article}

\usepackage[preprint, nonatbib]{neurips_2022}

\usepackage[utf8]{inputenc} 
\usepackage[T1]{fontenc}    
\usepackage{hyperref}       
\usepackage{url}            
\usepackage{booktabs}       
\usepackage{amsfonts}       
\usepackage{nicefrac}       
\usepackage{microtype}      
\usepackage{xcolor}         
\usepackage{csvsimple}
\usepackage{amsmath,amssymb,amstext} 
\usepackage{graphicx}

\title{GHN-Q: Parameter Prediction for Unseen Quantized Convolutional Architectures via Graph Hypernetworks}

\author{%
  Stone Yun\\
  Vision and Image Processing Group, University of Waterloo\\
  Waterloo Artificial Intelligence Institute\\
  \texttt{s22yun@uwaterloo.ca} \\
  \And
  Alexander Wong\\
  Vision and Image Processing Group, University of Waterloo\\
  Waterloo Artificial Intelligence Institute\\
  \texttt{a28wong@uwaterloo.ca} \\
}

\begin{document}

\maketitle

\begin{abstract}
  Deep convolutional neural network (CNN) training via iterative optimization has had incredible success in finding optimal parameters. However, modern CNN architectures often contain millions of parameters. Thus, any given model for a single architecture resides in a massive parameter space. Models with similar loss could have drastically different characteristics such as adversarial robustness, generalizability, and quantization robustness. For deep learning on the edge, quantization robustness is often crucial. Finding a model that is quantization-robust can sometimes require significant efforts. Recent works using Graph Hypernetworks (GHN) have shown remarkable performance predicting high-performant parameters of varying CNN architectures. Inspired by these successes, we wonder if the graph representations of GHN-2 can be leveraged to predict quantization-robust parameters as well, which we call GHN-Q. We conduct the first-ever study exploring the use of graph hypernetworks for predicting parameters of unseen quantized CNN architectures. We focus on a reduced CNN search space and find that GHN-Q can in fact predict quantization-robust parameters for various 8-bit quantized CNNs. Decent quantized accuracies are observed even with 4-bit quantization despite GHN-Q not being trained on it. Quantized finetuning of GHN-Q at lower bitwidths may bring further improvements and is currently being explored.
\end{abstract}

\section{Introduction}
\label{sec:intro}
AI-on-the-edge is an exciting area as we move towards an increasingly connected, but mobile world. However, there are tight constraints on latency, power, and area when enabling deep neural networks (DNN) for the edge. Consequently, fixed-precision, integer quantization such as in~\cite{TFQuant} has become an essential tool for fast, efficient CNNs.

Developing efficient models that can be deployed for low-power, quantized inference while retaining close to original floating point accuracy is a challenging task. In some cases, state-of-the-art performance can have significant degradation after quantization of weights and activations~\cite{MobileNetsQuantizePoorly}. So how can we find high-performant, quantization-robust parameters for a given CNN? 

Recent works by~\cite{PPUDA}~and~\cite{GHN1} have shown remarkable performance using Graph Hypernetworks (GHN) to predict all trainable parameters of unseen DNNs in a \textit{single forward pass}. For example, \cite{PPUDA} report that their GHN-2 can predict the parameters of an unseen ResNet-50 to achieve 60\% accuracy on CIFAR-10. Inspired by these successes, we wonder if the graph representational power of GHN-2 can be leveraged to predict quantization-robust parameters for unseen CNN architectures. We present the first-ever study exploring the use of GHNs to predict the parameters of unseen, quantized CNN architectures, which we call GHN-Q. By finetuning GHNs on a mobile-friendly CNN architecture space, we explore the use of adapting GHNs specifically to target efficient, low-power quantized CNNs. We find that even a floating point-finetuned GHN-Q can indeed predict quantization-robust parameters for various CNNs.

\begin{figure*}

\centerline{\includegraphics[width=12.5cm]{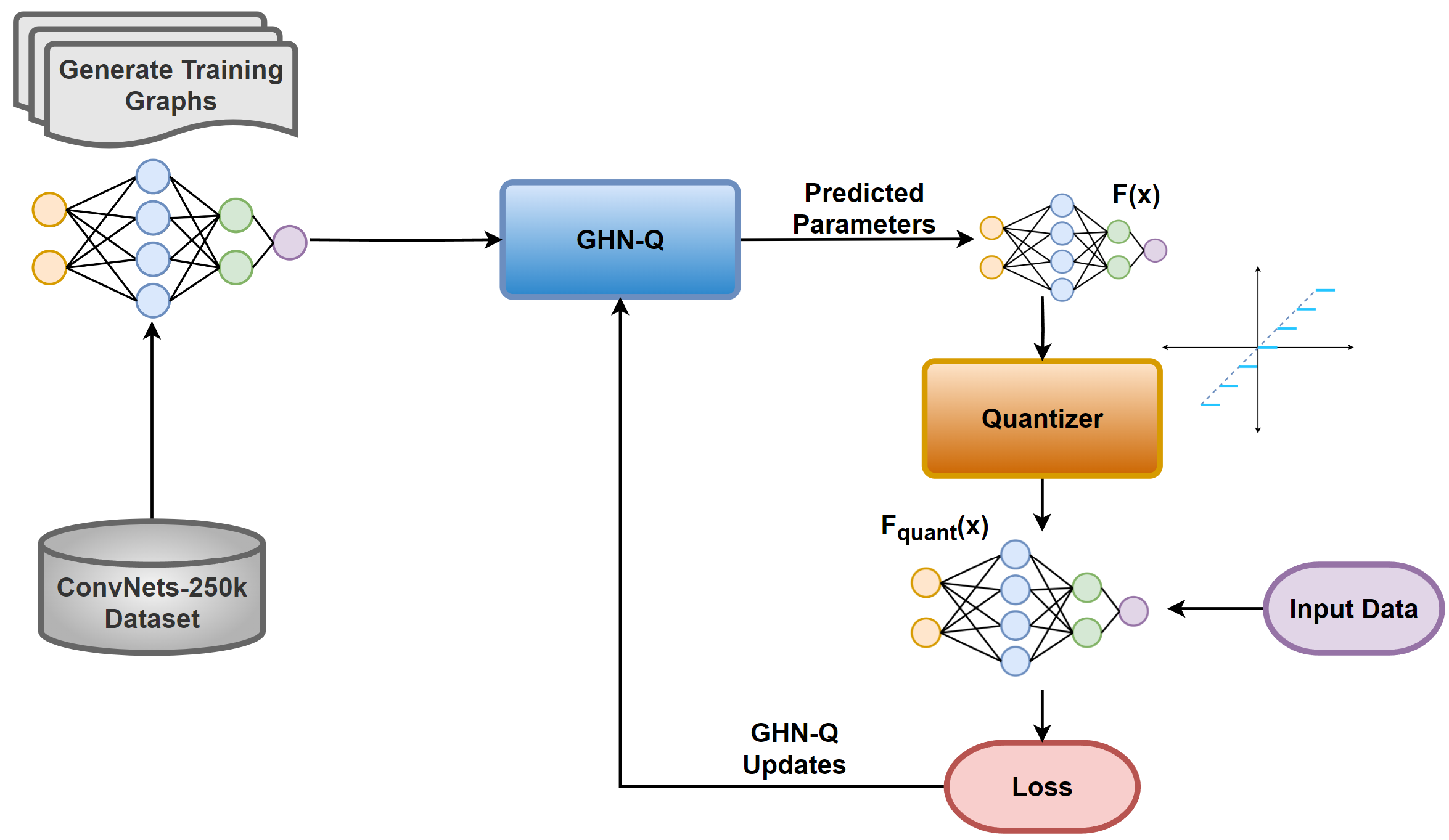}}
\caption[ghnq\_training]{Finetuning GHN-Q on ConvNets-250K. We generate a large number of CNN graphs which are then quantized to target bitwidth for training. Once trained, GHN-Q can predict robust parameters for unseen CNNs.}
\label{fig:GHNQ_train}
\end{figure*}

\section{Experiment}
\label{sec:experiment}
\vspace{-0.2in}
\begin{table*}[h]
    \caption{Testing GHN-Q on unseen quantized networks. CIFAR-10 top-1 and top-5 test accuracy of quantized CNNs (W8/A8, W4/A8, W4/A4, W2/A2) are compared to their full-precision (Float32) accuracy. Presented as (Mean$\%\pm$standard error of mean; Max$\%$). \textbf{ID} indicates in-distribution graphs sampled the same way as training set. \textbf{OOD} are out-of-distribution graphs with characteristics very different from those sampled in training.}
    
    \centering
    \begin{tabular}{c| c| ccc}
    \toprule
     \textbf{Top-1 Accuracy}& \textbf{ID} & \multicolumn{3}{c}{\textbf{OOD}}\\
     \textbf{by Bitwidth} & Test & Deep &  Wide &  BN-Free \\
     \midrule
      Float32 & $71.1\pm0.3; 80.2$ & $68.1\pm0.7; 79.8$ & $69.8\pm0.5; 79.0$ & $37.8\pm1.3; 56.1$ \\
      W8/A8 & $70.9\pm0.3; 80.1$ & $67.9\pm0.7; 79.5$ & $69.6\pm0.5; 78.6$ & $37.5\pm1.3; 56.4$ \\
      W4/A8 & $47.4\pm 0.4; 63.9$ & $39.0\pm 0.7; 62.9$ & $43.8\pm 0.6; 63.3$ & $25.7\pm 0.9; 43.7$\\
      W4/A4 & $37.2\pm0.3; 52.6$ & $30.7\pm0.5; 50.2$ & $34.7\pm0.5; 50.7$ & $21.5\pm0.8; 36.7$\\
      W2/A2 & $10.0\pm0.0; 11.0$ & $10.0\pm0.0; 10.7$ & $10.0\pm0.0; 10.7$ & $9.7\pm0.0; 10.4$\\
      \bottomrule
      \textbf{Top-5 Accuracy} & \textbf{ID} & \multicolumn{3}{c}{\textbf{OOD}}\\
     \textbf{by Bitwidth} & Test & Deep &  Wide &  BN-Free \\
     \midrule
      Float32 & $97.3\pm0.1; 99.0$ & $96.4\pm0.2; 98.8$ & $97.2\pm0.1; 98.7$ & $84.3\pm1.2; 94.8$ \\
      W8/A8 & $97.3\pm0.1; 98.9$ & $96.4\pm0.2; 98.8$ & $97.2\pm0.1; 98.7$ & $84.1\pm1.2; 94.9$ \\
      W4/A8 & $89.8\pm 0.2; 95.9$ & $85.2\pm 0.4; 95.8$ & $88.0\pm 0.3; 95.5$ & $74.6\pm 1.2; 89.8$\\
      W4/A4 & $83.7\pm0.2; 91.7$ & $78.6\pm0.4; 91.0$ & $82.1\pm0.3; 91.4$ & $69.3\pm1.1; 87.0$\\
      W2/A2 & $50.0\pm0.0; 51.4$ &  $50.0\pm0.0; 50.9$ & $50.0\pm0.0; 51.4$ & $50.2\pm0.1; 52.0$\\
      \bottomrule
    \end{tabular}
    \label{tab:results}
\end{table*}

We would first like to investigate whether full precision floating point training on a target design space can train GHN-Q to predict high-performant CNN parameters that are robust to 8-bit uniform quantization. A couple aspects of the method in~\cite{PPUDA} suggest that the parameters predicted by GHN-Q should be compact and quantization-friendly, namely the channel-wise weight tiling and differentiable parameter normalization. We finetuned a CIFAR-10, DeepNets-1M pretrained GHN-2 model obtained from~\cite{PPUDA_Github} on a set of $2.5\times10^5$ mobile-friendly architectures that we call ConvNets-250K. Figure~\ref{fig:GHNQ_train} shows how we train GHN-Q for predicting quantization-robust CNN parameters.

The ConvNets-250K search space consists solely of convolution\footnote{Various convolutions such as depthwise, dilated, regular.} (including residual blocks), batch-normalization (BatchNorm), pooling, and linear layers as these are typically the easiest to accelerate on edge devices. We also limit the maximum number of parameters in sampled CNNs to $10^7$. We could have set a lower constraint but we also wanted to allow for a more diverse search space. Finetuning on ConvNets-250K was run for 100 epochs using CIFAR-10~\cite{cifar10}. Initial learning rate was $0.001$ and reduced by a factor of $0.1$ at epoch 75. GHN-Q is trained with Adam optimizer using $\beta_1 = 0.9$, $\beta_2 = 0.999$, weight decay of $10^{-5}$, training batch-size of 32, and meta-batch-size of 4. We continue to use the weight-tiling and parameter normalization described in~\cite{PPUDA} and we use $s^{(max)}=10$ as the maximum shortest path for virtual edges.

While we start with full-precision finetuning of GHN-Q on the target design space, we can easily quantize the CNNs with arbitrary precision or model other scalar quantization methods. Thus, GHN-Q becomes a powerful tool for quantization-aware design of efficient CNN architectures.

\section{Results}
\label{sec:results}

We follow a testing procedure similar to~\cite{PPUDA} and evaluate the trained GHN-Q by comparing the mean CIFAR-10 test accuracy at precisions of 32-bit floating point (Float32), 8-bit quantization (W8/A8), and 4-bit quantization\footnote{We use tensorwise, asymmetric, uniform quantization throughout.} (W4/A4) where W denotes weights bitwidth and A denotes activations bitwidth. Additionally, we also tested W4/A8 and W2/A2 settings. Table~\ref{tab:results} shows  the results on our different testing splits. BN-Free networks have no BatchNorm layers. Wide/Deep indicate \textbf{much} wider/deeper nets than those seen in training. Thus, they are labelled as out-of-distribution (OOD).

For handling BatchNorm, we use a test-batch-size of 64 to get batch statistics~\cite{BatchNorm}. We also have to recompute BatchNorm-folded (BN-Fold) weights each batch before quantizing the BN-Fold weights like in~\cite{TFQuant}. See Eq.~\ref{eq:bnfold} for the BN-Fold operation where $\gamma$ is a learnable parameter, $\sigma^2_B$ is batch variance (could also be exponential moving average of variance) and $\epsilon$ is a small constant. Quantization encodings use the absolute tensor ranges.

\begin{equation}
\label{eq:bnfold}
w_{fold} = \frac{\gamma w}{\sqrt{\sigma^2_B + \epsilon}}
\end{equation}

It is worth noting that we only quantize the weights and activations\footnote{Note that we did not yet add $Quantize()$ after concatenation.} with a $Quantize()$ operator instead of running fully fixed-point inference. Rounding and truncation errors of fully fixed-point arithmetic will lead to some additional error. However, as most of the quantization noise is due to weights and activations, the simulated quantization generally correlates well with on-device accuracy~\cite{TFQuant, TFQuantWhitePaper, AIMET}.

\section{Discussion}
\label{sec:discussion}
The parameters predicted by GHN-Q are surprisingly robust despite not having been trained on any kind of quantization. It is particularly interesting that even for 4-bit quantization, the average test accuracy is significantly better than random chance. A likely explanation is that the channel-wise weight tiling and differentiable parameter normalization lead to layerwise distributions that are compact and quantization-friendly. In~\cite{MobileNetsQuantizePoorly}, the authors find that a mismatch between channelwise distributions can lead to significant accuracy loss for post-training quantization. The weight tiling method in~\cite{PPUDA} copies predicted parameters across channels and thus minimizes such distributional mismatch by-construction. Additionally, parameter normalization could help produce less heavy-tailed distributions. A detailed analysis of the distributions of GHN-Q predicted parameters would yield a clearer picture. As depthwise-separable convolution is particularly susceptible to distributional mismatch, it would be interesting to test the 8-bit quantized performance of GHN-Q on a test set consisting solely of MobileNet-like CNNs such as those in~\cite{MobileNetV1, MobileNetV2}.

Besides analyzing mean quantized accuracy, quantization robustness needs to be quantified on a per-network basis. An analysis of the mean accuracy change and quantization error (e.g., quantized mean squared error) of individual networks would provide better insight. It would be interesting to see how quantization error may change after 8-bit quantized GHN-Q finetuning even if accuracy remains similar.

We believe these results demonstrate great potential for leveraging the powerful graph representation of GHNs for edge-AI. Finetuning GHN-Q on lower bitwidths such as 4-bit and 2-bit quantized networks should further improve quantization robustness of predicted parameters. Furthermore, GHN-Q could possibly be a useful weight initialization for quantized CNN training. Quantized models usually require full-precision training to convergence before quantization-aware finetuning. In~\cite{PPUDA} there are questions of how well GHN-2 predicted parameters can be used for finetuning on the source task. However, if GHN-Q could be adapted such that predicted parameters can directly start quantization-aware training, there would be significant savings in training CNNs for quantization.

\bibliographystyle{IEEEtran}
\bibliography{main}
\end{document}